\documentclass{article} 
\usepackage{arxiv,times}

\usepackage[utf8]{inputenc} 
\usepackage[T1]{fontenc}    
\usepackage{hyperref}       
\usepackage{url}            
\usepackage{booktabs}       
\usepackage{amsfonts}       
\usepackage{nicefrac}       
\usepackage{microtype}      
\usepackage{amssymb}
\usepackage{subcaption}
\usepackage{epigraph}
\usepackage{array}
\usepackage{multirow}
\usepackage[linesnumbered,ruled]{algorithm2e}
\usepackage{algpseudocode}
\usepackage{amsmath,amssymb,mathtools}
\usepackage{cuted}
\usepackage{amsthm,bm}
\usepackage{afterpage}
\usepackage{bbm}
\usepackage{color, colortbl}
\usepackage{color}
\usepackage{algorithm2e}
\usepackage{mathtools}
\usepackage{float}
\usepackage{wrapfig}

\newcommand{\V}[1]{\mathbf{#1}}

\newcommand{\bfx}{\V{x}}

\newcommand{\xad}{\bfx_{\text{adv}}}

\theoremstyle{plain}
\newtheorem{Thm}{Theorem}

\newtheorem{Df}{Definition}

\DeclareMathOperator*{\argmin}{arg\,min}
\usepackage{bbm}

\newif\ifsubmit

\submittrue

\ifsubmit
\newcommand{\gerald}[1]{}
\newcommand{\bo}[1]{}
\newcommand{\ruoxi}[1]{}
\newcommand{\jingkang}[1]{}
\newcommand{\dawn}[1]{}
\else
\newcommand{\gerald}[1]{{\color{magenta}[Gerald: #1]}}
\newcommand{\bo}[1]{{\color{blue}[Bo: #1]}}
\newcommand{\ruoxi}[1]{{\color{cyan}[Ruoxi: #1]}}
\newcommand{\jingkang}[1]{{\color{orange}[Jingkang: #1]}}
\newcommand{\dawn}[1]{{\color{red}[Dawn: #1]}}
\fi

\title{One Bit Matters: Understanding Adversarial Examples as the Abuse of Redundancy}

\author{Jingkang Wang \\
Shanghai Jiao Tong University\\
Shanghai, China\\
\texttt{wangjksjtu@gmail.com}
\And
Ruoxi Jia \\
University of California, Berkeley \hspace{1.22cm} \\
California, USA\\
\texttt{ruoxijia@berkeley.edu} \\
\AND
Gerald Friedland \\
Lawrence Livermore National Labs \& \\
University of California, Berkeley\\
California, USA\\
\texttt{fractor@eecs.berkeley.edu}
\And
Bo Li \\
University of Illinois at Urbana-Champaign \\
Illinois, USA \\
\texttt{lxbosky@gmail.com}
\And
Costas Spanos \\
University of California, Berkeley \\
California, USA\\
\texttt{spanos@berkeley.edu}
}

\iclrfinalcopy 

\begin{document}

\maketitle

\begin{abstract}
\ruoxi{Comments on the title: strictly speaking, we show that redundancy in *features* lead to adversarial examples, rather than redandancy in *data*. So we may consider to modify the title in order to abide by our foundings}
\bo{how about information redundancies?}
\gerald{New title OK? The term features is loaded in too many ways. Shorter is usually better}
Despite the great success achieved in machine learning (ML), adversarial examples have caused concerns with regards to its trustworthiness: A small perturbation of an input results in an arbitrary failure of an otherwise seemingly well-trained ML model. While studies are being conducted to discover the intrinsic properties of adversarial examples, such as their transferability and universality, there is insufficient theoretic analysis to help understand the phenomenon in a way that can influence the design process of ML experiments. In this paper, we deduce an information-theoretic model which explains adversarial attacks as the abuse of feature redundancies
in ML algorithms. We prove that feature redundancy is a necessary condition for the existence of adversarial examples. Our model helps to explain some major questions raised in many anecdotal studies on adversarial examples.
Our theory is backed up by empirical measurements of the information content of benign and adversarial examples on both image and text datasets. Our measurements show that typical adversarial examples introduce just enough redundancy to overflow the decision making of an ML model trained on corresponding benign examples. We conclude with actionable recommendations to improve the robustness of machine learners against adversarial examples.
\end{abstract}


\section{Introduction}
Deep neural networks (DNNs) have been widely applied to various applications and achieved great successes~\citep{ciregan2012multi,hinton2012deep,sallab2017deep,Chen2018}. This is mostly due to their versatility: DNNs are able to be trained to fit any specific target functions. Therefore, it raises great concerns given the discovery that DNNs are vulnerable to \emph{adversarial examples}. These are carefully crafted inputs, which are often seemingly normal within the variance of the training data but can fool a well-trained model with high attack success rate~\citep{goodfellow2014explaining}. Adversarial examples can be generated for various types of data, including images, text, audio, and software~\citep{carlini2018audio,ebrahimi2017hotflip}, and for different ML models, such as classifiers, segmentation models, object detectors, and reinforcement learning~\citep{kos2017delving,huang2017adversarial}.
Moreover, adversarial examples are transferable~\citep{tramer2017space,liu2016delving} --- if we generate adversarial perturbation against one model for a given input, the same perturbation will have high probability to be able to attack other models trained on similar data, regardless how different the models are. 
Last but not the least, adversarial examples cannot only be synthesized in the digital world but also in the physical world~\citep{evtimov2017robust,kurakin2016adversarial}, which has caused great real-world security concerns. 

Given such subtle, yet universally powerful attacks against ML models, several defensive methods have been proposed. For example, \cite{liu2018feature,feinman2017detecting} pre-process inputs to eliminate certain perturbations. Other work~\citep{cao2017mitigating} suggest to push the adversarial instance into random directions so they hopefully escape a local minimum and fall back to the correct class. 
There are work to establish metrics to distinguish adversarial examples from benign ones so that one can filter out adversarial examples before they are used by ML models~\citep{li2017adversarial}.
However, so far, all defense and detection methods have shown to be adaptively attackable~\citep{athalye2018obfuscated,carlini2017adversarial}. Therefore, intelligent attacks against intelligent defenses become an arms race. Defending against adversarial examples remains an open problem. 


In this paper, we propose and validate a theoretical model that can be used to create an actionable understanding of adversarial perturbations. Based upon the model, we give recommendations to modify the design process of ML experiments such that the effect of adversarial attacks is mitigated. We illustrate adversarial examples using an example of a simple perceptron network that learns the Boolean equal operator and then generalize the example into a universal model of classification based on Shannon's theory of communication. We further explain how adversarial examples fit the thermodynamics of computation. We prove a necessary condition for the existence of adversarial examples. 
In summary, the contributions of the paper are listed below:
\begin{itemize}
    \item We provide an explanation for adversarial examples from the perspective of information theory and thermodynamics of computing~\citep{feynman88}, and show it is consistent with existing observations;
    \item We theoretically prove that information redundancy is a necessary condition for the vulnerability of ML models to adversarial examples;
    \item We conduct extensive experiments that showcase the relationship between feature redundancy and adversarial examples
    \item We provide actionable recommendations for potentially mitigating adversarial effects within machine learning systems.
\end{itemize}

\section{Related Work}
\label{sec:related}
Given a benign sample $\bfx$, an adversarial example $\xad$ is generated by adding a small perturbation $\epsilon$ to $\bfx$ (i.e. $\xad = \bfx + \epsilon$), so that $\xad$ is misclassified by the targeted classifier $g$. Related work has mostly focused on describing the properties of adversarial examples as well as on defense and detection algorithms. 

Goodfellow et al. have hypothesized that the existence of adversarial examples is due to the linearity of DNNs~\citep{goodfellow2014explaining}. Later, boundary-based analysis has been derived to show that adversarial examples try to cross the decision boundaries~\citep{he2018decision}. More studies regarding to data manifold have also been leveraged to better understand these perturbations~\citep{ma2018characterizing,gilmer2018adversarial,wang2016theoretical}. While these works provide hints to obtain a more fundamental understanding, to the best of our knowledge, no study was able to create a model that results in actionable recommendations to improve the robustness of machine learners against adversarial attacks. Prior work does not suggest a measurement process or theoretically show the necessary or sufficient conditions for the existence of adversarial examples.

Several approaches have been proposed to generate adversarial examples. For instance, the fast gradient sign method has been proposed to add perturbations along the gradient directions~\citep{goodfellow2014explaining}. Other examples are optimization algorithms that search for the minimal perturbation~\citep{carlini2017towards,liu2016delving}. Based on the adversarial goal, attacks can be classified into two categories: targeted and untargeted attacks.  In a targeted attack, the adversary's objective is to modify an input $\bfx$ such that the target model $g$ classifies the perturbed input $\xad$ as a \emph{targeted} class chosen, which differs from its ground truth. In a untargeted attack, the adversary's objective is to cause the perturbed input $\xad$ to be misclassified in \emph{any class} other than its ground truth. Based on the adversarial capabilities, these attacks can be categorized as white-box and black-box attacks; an adversary has full knowledge of the classifier and training data in the white-box setting~\citep{szegedy2014intriguing,goodfellow2014explaining,carlini2016towards,moosavi2015deepfool,papernot2016limitations,biggio2013evasion,fawzi2015manitest,kanbak2017measuring,kurakin2016adversarial}, but zero knowledge about them in the black-box setting~\citep{papernot2016transferability,liu2016delving,moosavi2016universal,mopuri2017fast}.

Interestingly enough, adversarial examples are not restricted to ML. Intuitively speaking, and consistent with the model presented in this paper, acoustic noise masking could be regarded as an adversarial attack on our hearing system. Acoustic masking happens, for example, when a clear sinusoid tone cannot be perceived anymore because a small amount of white noise has been added to the signal~\citep{phillips1990neural}. This effect is exploited in MP3 audio compression and privacy applications. Similar examples exist, such as optical illusions in the visual domain~\citep{ponomarenko2007between} and defense mechanisms against sensor-guided attacks~\citep{warm1997apparatus} in the military domain.

\section{A Model for Adversarial Examples}
\begin{figure}
  \centering
  \resizebox{0.9\columnwidth}{!}{
  \begin{subfigure}[b]{0.49\textwidth}
  \includegraphics[width=\textwidth]{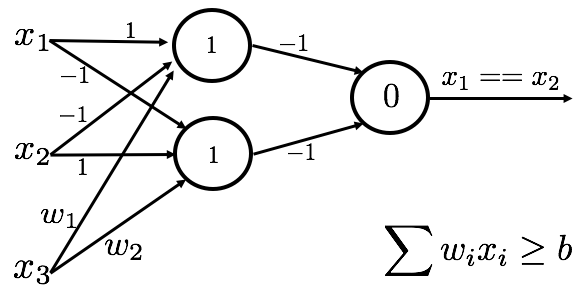}
  \caption{ 
  }
  \label{fig:nxornet}
  \end{subfigure}
  ~
  \begin{subfigure}[b]{0.49\textwidth}
  \includegraphics[width=\textwidth]{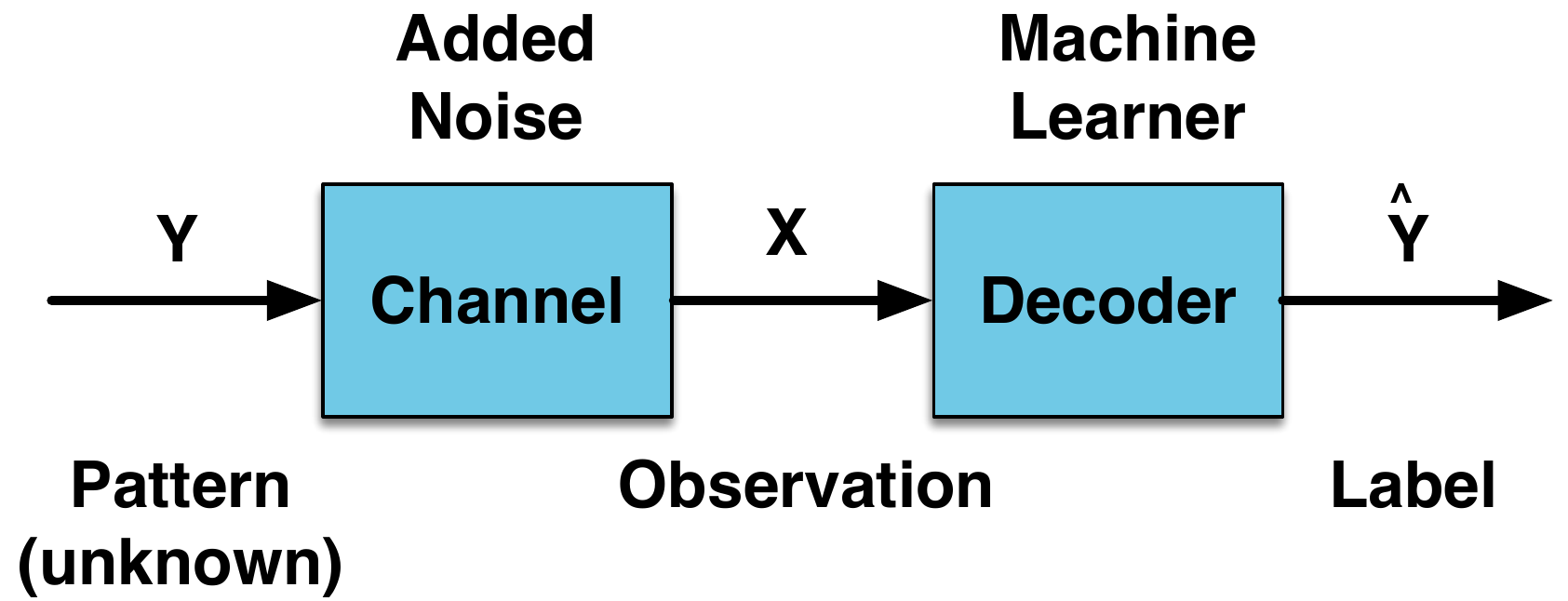}
  \caption{
 }
  \label{fig:shannonmodel}
  \end{subfigure}
  }
  \caption{(a) A perceptron network implementing the Boolean equality of $x_1$ and $x_2$ with added redundancy $x_3$ (see Section~\protect\ref{sec:boolmodel}). (b) Shannon channel model of machine learning inspired by \protect\cite{mackay2002}. (see Section~\protect\ref{sec:shannonmodel}).}
\end{figure}

Intuitively speaking, we want to explain the phenomenon shown in Figure~\ref{fig:xornoise}, which depicts a plane filled with points that were originally perfectly separable with two 2D linear separations. As the result of perturbing several points by a mere 10\% of the original position, the separation of the two classes requires many more than two linear separators. This shows a small amount of noise can overflow the separation capability of a network dramatically. In the following section, we introduce an example model along which we will derive our mathematical understanding, consistent with our experiments in Section~\ref{sec:experiments} and the related work mentioned in Section~\ref{sec:related}. 

\subsection{Adversarial Examples and Boolean Functions}
\label{sec:boolmodel}

Consider a perceptron network which implements the Boolean equal function ("NXOR") between the two variables $x_1$ and $x_2$. The input $x_3$ is redundant in the sense that the result of $x_1==x_2$ is not influenced by the value of $x_3$. 

The first obvious observation is that adding $x_3$ doubles the input space. Instead of $2^2=4$ possible input pairs, we now have $2^3=8$ possible input triples that the network needs to map to sustain the result $x_1==x_2$ for all possible combinations of $x_1$, $x_2$, $x_3$. The network architecture shown in Figure~\ref{fig:nxornet}, for example, theoretically has the capacity to be trained to learn all $8$ input triples~\citep{friedlandkrell2017,friedland2018}. Translating this example into a practical ML scenario, however, this would mean that we have to exhaustively train the entire input space for all possible settings of the noise. This is obviously unfeasible.    

\begin{figure}[h]
  \centering
  \resizebox{0.65\columnwidth}{!}{
  \begin{subfigure}[b]{0.3\textwidth}
 \includegraphics[width=\textwidth]{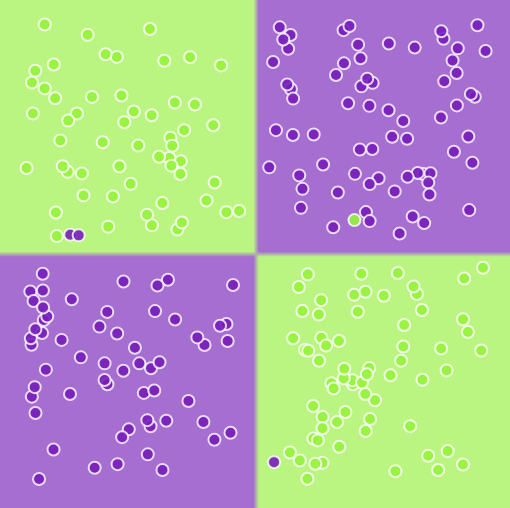}
  \caption{}
  \label{fig:xornoise}
  \end{subfigure}
  \hspace{0.5cm}
  \begin{subfigure}[b]{0.45\textwidth}
  \begin{tabular}{lll}
    \toprule
    \#Adv. Examples & \#Allowing Edges & Potential \\
    \midrule
    0 & 0 & 0 \\
    4 & 1 & 1 \\
    4 & 1 & 1 \\
    4 & 2 & 0 \\
    4 & 1 & 1 \\
    4 & 1 & 1 \\
    6 & 2 & 0 \\
    6 & 2 & 2 \\
    6 & 2 & 2 \\
    \bottomrule
  \end{tabular}
  \caption{}
  \label{table:xorsuppress}
  \end{subfigure}
 }
 \caption{(a) A network trained on XOR cannot separate the two classes anymore perfectly once we add 10\% of noise to some of the inputs. (b) Suppression capability of adversarial attacks against the Boolean equality network.}
\end{figure}

We will therefore continue our analysis in a more practical setting. We assume a network like in Figure~\ref{fig:nxornet} is correctly trained to model $x_1==x_2$ in the absence of a third input. One example configuration is shown.  Now, we train weights $w_1$ and $w_2$ to try to suppress the redundant input $x_3$ by going through all possible combinations for $w_i\in \{-1,0,1\}$. This weight choice is without losing generality as the inputs $x_i$ are $\in \{0,1\}$ (see~\cite{Rojas1992}). An adversarial example is defined as a triple $(x_1,x_2,x_3)$ such that the output of the network is not the result of $x_1==x_2$. Simulating through all configurations exhaustively results in Table~\ref{table:xorsuppress}. The only case that allows for $100$\% accuracy, i.e., no adversarial examples, is the setting $w_1=w_2=0$, in which case $x_3$ is suppressed completely. In the other cases, we can roughly say that the more the network pays attention to $x_3$, the worse the result (allowing edges). This shows the result is better if one of the $w_i$ is set to $0$ compared to none. Furthermore, the higher the potential, defined as the difference between the maximum and the minimum possible activation value as scaled by the $w_i$, the worse the result is. The intuition behind this is that higher potential change leads to higher potential impacts to the overall network. 

Using this simple model, one can see the importance of suppressing noise. Thresholds of neurons taking redundant inputs should be high, or equivalently, weights should be close to 0 (and equal to 0 in the optimal scenario). Now generalizing the example to a large network training images with 'real-valued' weights, it becomes clear that redundant bits of an image should be suppressed by low enough weights otherwise it is easy to generate an exponential explosion of patterns needed to be recognized. 


\subsection{General Model}
\label{sec:shannonmodel}
The generalization of the example from the previous section is shown in Figure~\ref{fig:shannonmodel}. The model shows a machine learner performing the task of matching an unknown noisy pattern to a known pattern (label). For example, a perceptron network implements a function of noisy input data. It quantizes the input to match a known pattern and then outputs the results of the learned function from a known pattern to a known output. Formally, the random variable $Y$ encodes unknown patterns that are sent over a noisy channel. The observation at the output of the channel is denoted by the random variable $X$. For example, $X$ could represent image pixels.
The machine learner then erases all the noise bits in $X$ to match against trained patterns which are then mapped to known outputs $\hat{Y}$, for example, the labels. It is well known from the thermodynamics of computing~\citep{feynman88} that setting memory bits and copying them is theoretically energy agnostic. However, resetting bits to zero is not. In other words, we need to spend energy (computation) to reset the noisy bits added by the channel and captured in the observation to get to a distribution of patterns $\hat{Y}$ that is isomorphic to the original (unknown) distribution of patterns $Y$. Connecting back to the NXOR example from the previous section, $Y$ would be the distribution over the input variables $x_1$ and $x_2$. The noise added is modeled by $x_3$ and $\hat{Y}$ is the desired output isomorphic to $x_1$ and $x_2$ being equal. Now assuming a fully trained model, this model allows us to explain several phenomena explored in the introduction and Section~\ref{sec:related}. 

First, as illustrated in the previous section, we can view the machine learner as a trained bit eraser. The machine learner has been trained to erase exactly those bits that are irrelevant to the pattern to be matched. This elimination of irrelevance constitutes the generalization capability. For a black box adversarial attack, we therefore just need to add enough irrelevant input to overflow this bit erasure function. As a result, insufficient redundant bits can be absorbed and the remaining bits now create an exponential explosion for the pattern matching functionality. In a whitebox attack, an attacker can guess and check against the bit erasing patterns of the trained machine learner and create a sequence of input bits that specifically overflows the decision making. In both cases, our model predicts that adversarial patterns should be harder to learn as they consist of more bits to erase. This is confirmed in our experiments in Section~\ref{sec:experiments}. It is also clear that the theoretical minimum overflow is one bit, which means, small perturbations can have big effects. This will be made rigorous in Section~\ref{sec:proof}. It is also well known that, for example, in the image domain one bit of difference is not perceivable by a human eye. Training with noisy examples will most likely make the machine learner more robust as it will learn to reduce redundancies better. However, a specific whitebox attack (with lower entropy than random noise), which constitutes a specific perceptron threshold overflow, will always be possible because training against the entire input space is unfeasible. 

Second, with training data available, it is highly likely that a surrogate machine learner will learn to erase the same bits. This means that similar bit overflows will work on both the surrogate and the original ML attack, thus explaining transferability-based attacks. 


\subsection{Abuse of Redundancy}
\label{sec:proof}
In the following we will present a proof based on the model presented in Section~\ref{sec:shannonmodel} and the currently accepted definition of adversarial examples~\citep{wang2016theoretical} that shows that feature redundancy is indeed a necessary condition for adversarial examples. Throughout, we assume that a learning model can be expressed as  $f(\cdot)=g(T(\cdot))$, where $T(\cdot)$ represents the feature extraction function and $g(\cdot)$ is a simple decision making function, e.g., logistic regression, using the extracted features as the input.

\begin{Df}[Adversarial example~\citep{wang2016theoretical}]
Given an ML model $f(\cdot)$ and a small perturbation $\delta$, we call $x'$ an adversarial example if there exists $x$, an example drawn from the benign data distribution, such that $f(x) \neq f(x')$ and $\|x-x'\|\leq \delta$.
\end{Df}

We first observe that $\forall x, x'~\exists \delta$ such that $\|x-x'\|\leq \delta \implies f(x)=f(x')$ is the generalization assumption of a machine learner. The existence of an adversarial $x'$ is therefore equivalent to a contradiction of the generalization assumption. This is, $x'$ could be called a counter example. Practically speaking, a single counter example to the generalization assumption does not make the machine learner useless though. In the following, and as explained in previous sections, we connect the existence of adversarial examples to the information content of features used for making predictions.

\begin{Df}[Feature redundancy]
Let $X$ and $Y$ represent the random variables corresponding to features and unknown pattern, respectively. Let $T_Y^\text{min}(X)$ denote the minimal sufficient statistic of $X$, i.e., $T_Y^\text{min}(X) = \argmin_{T(X):I(X;Y) = I(T(X);Y)} H(T(X))$, where $T(X)$ is the sufficient statistic for $Y$. The redundancy of using $T(\cdot)$ as the feature extractor for predicting $Y$ is defined as
\begin{align}
    R(T(X);Y) = H(T(X)) - H(T_Y^\text{min}(X))
\end{align}
\end{Df}

\begin{Thm}
\label{thm:necessary}
Suppose that the feature extractor $T(X)$ is a sufficient statistic for $Y$ and that there exist adversarial examples for the ML model $f(\cdot)=g(T(\cdot))$, where $g(\cdot)$ is an arbitrary decision making function. 
%
Then, $T(X)$ is not a minimal sufficient statistic. 

\end{Thm}

We leave the proof to the appendix. The idea of the proof is to explicitly construct a feature extractor with lower entropy than $T(X)$ using the properties of adversarial examples. Theorem~\ref{thm:necessary} shows that the existence of adversarial examples implies that the feature representation contains redundancy. We would expect that more robust models will generate more succinct features for decision making. We will corroborate this intuition in Section~\ref{sec:output_complexity}.
 

\section{Experimental Results}
\label{sec:experiments}
In this section, we provide empirical results to justify our theoretical model for adversarial examples. Our experiments aim to answer the following questions. First, are adversarial examples indeed more complex (e.g. they contain more redundant bits with respect to the target that need to be erased by the machine learner)? If so, adversarial examples should require more parameters to memorize in a neural network. Second, is feature redundancy a large enough cause of the vulnerability of DNNs that we can observe it in a real-world experiment? Third, can we exploit the higher complexity of adversarial examples to possibly detect adversarial attacks? Fourth, does quantization of the input indeed not harm classification accuracy?

\subsection{Capacity Measurements}
\begin{figure}
  \centering
  \resizebox{0.85\columnwidth}{!}{
  \includegraphics[width=0.475\linewidth]{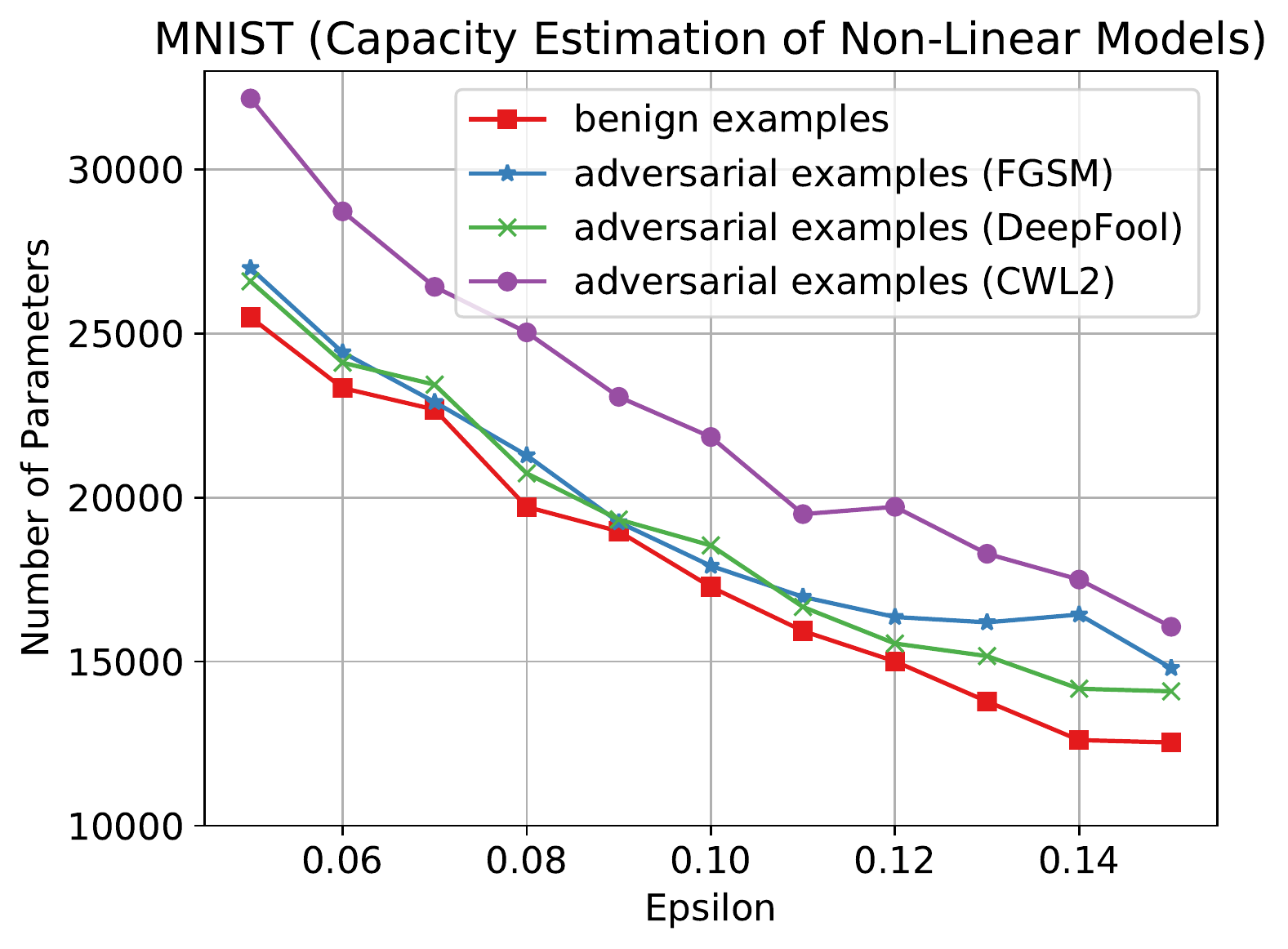}
  \includegraphics[width=0.45\linewidth]{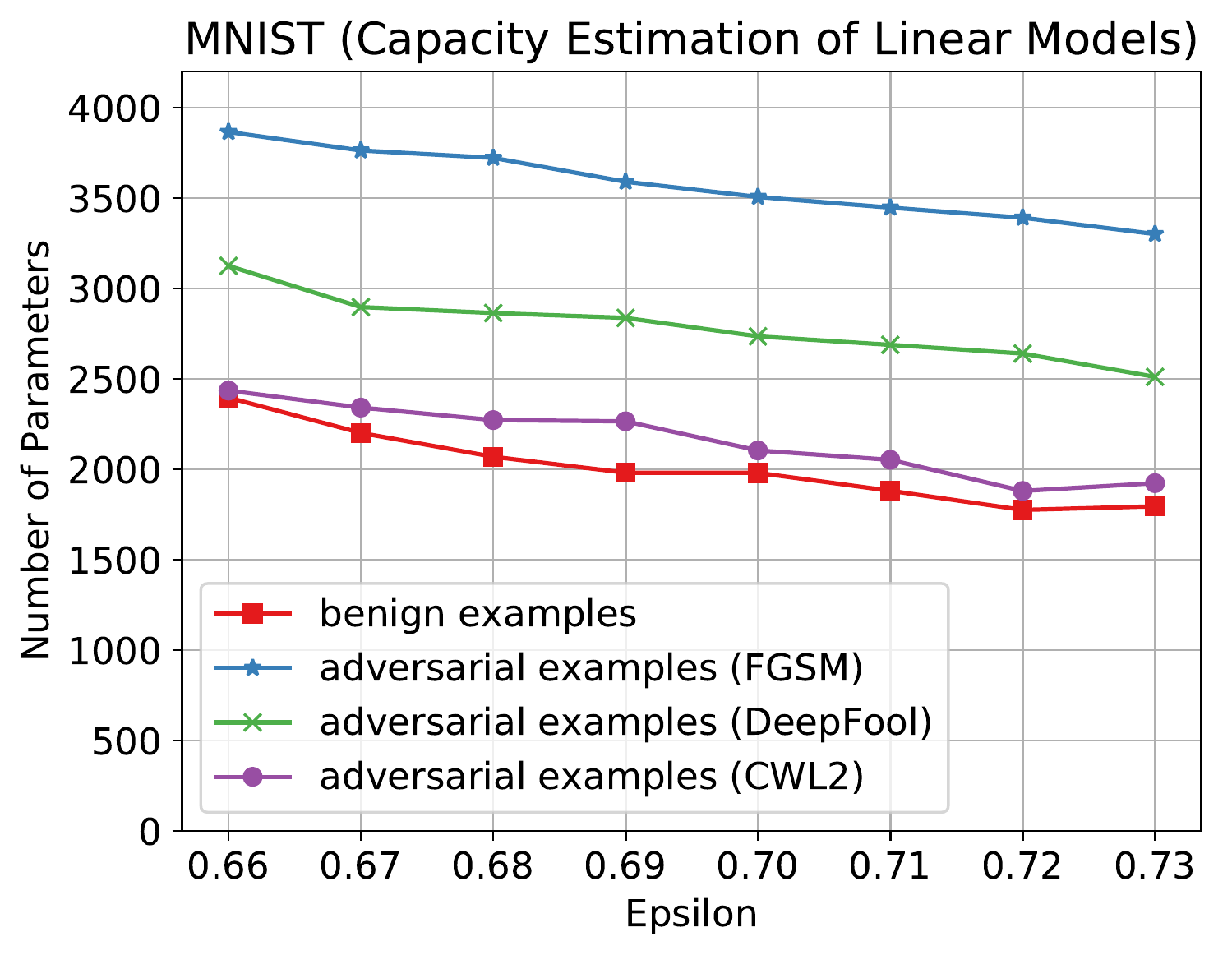}
  }
  \caption{Comparison of estimated model capacity (minimal number of parameters required) as a function of allowed error $\epsilon$ under different attack algorithms on MNIST. The results are obtained based on the MLPs with and without non-linear activation functions (e.g. ReLU). 
  Note that the non-linear models can generalize the data better, so the $\epsilon$ is fixed in a much smaller range.}
  \label{fig:cap}
\end{figure}

\begin{figure}
  \centering
  \includegraphics[width=0.98\linewidth]{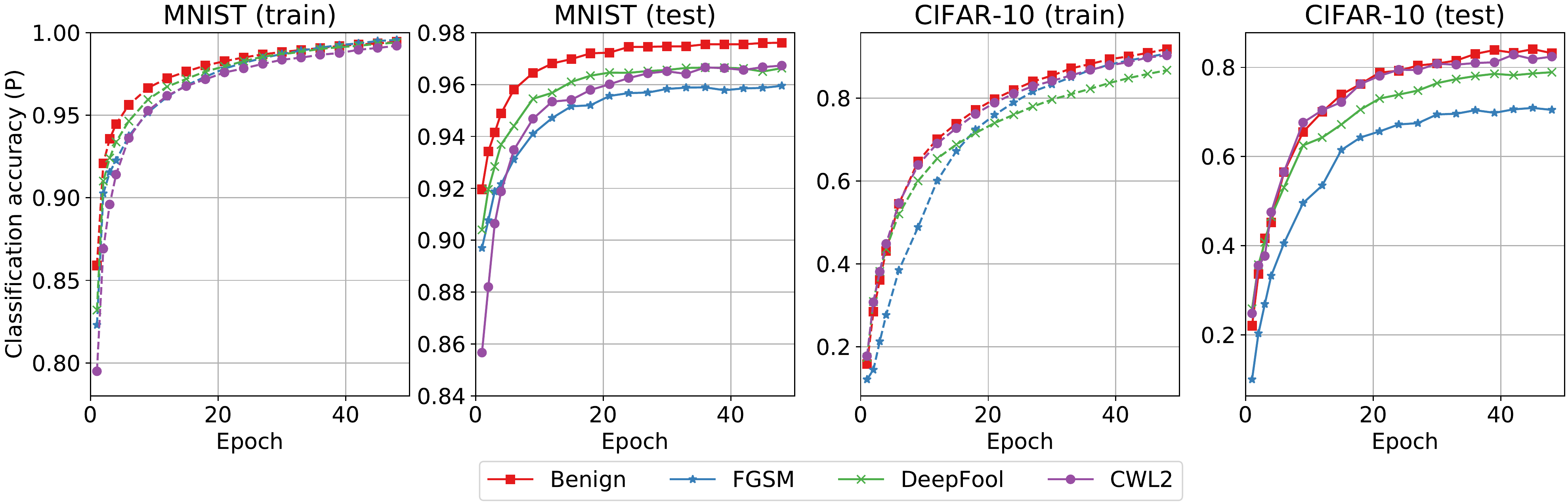}
  \caption{Classification accuracy of different attack models as a function of training epoch. Adversarial examples are generated with different attack methods on both MNIST and CIFAR-10.}
  \label{fig:acc}
  \vspace{-5mm}
\end{figure}

Our model implies that adversarial examples generally have higher complexity than benign examples. In order to evaluate this claim practically, we need to show that this complexity increase is in fact an increase of irrelevant bits with regards to the encoding performed in neural networks towards a target function. This can be established by showing that adversarial examples are more difficult to memorize than benign examples. In other words, a larger model capacity is required for training adversarial examples. To quantitatively measure how much extra capacity is needed, we measure the capacity of multi-layer perceptrons (MLP) models with or without non-linear activation function (ReLU) on MNIST. Here we define the model capacity as the minimal number of parameters needed to memorize all the training data. To explore the capacity, we first build an MLP model with one hidden layer (units: 64). This model is efficient enough to achieve high performance and memorize all training data (with ReLU). After that, weights are reduced by randomly setting some of their values to zero and marking them untrainable. The error $\epsilon$ is set to evaluate the training success (training accuracy is larger than $1-\epsilon$). We explore the minimal number of parameters and utilize binary search to reduce computation complexity. Finally, we change different $\epsilon$ and repeat the above steps. As illustrated in Figure~\ref{fig:cap}, the benign examples always require fewer number of weights to memorize on different datasets with various attack methods. It is shown that adversarial examples indeed require larger capacity. From the training/testing process given in Figure~\ref{fig:acc}, we can draw the same conclusion. The benign examples are always fitted and predicted more efficiently than adversarial examples given the same model. That is to say, adversarial examples have more complexity and therefore require higher model capacity. 


\definecolor{Gray}{gray}{0.87}
\begin{table}
\centering
\caption{Comparison of complexity for benign and adversarial examples on MNIST and CIFAR-10.}
\label{tab:entropy_img}
\begin{small}
\begin{tabular}{@{}cccccc@{}}
\toprule
Dataset                                        & Examples & H (MLE) & H (JVHW) & Original Size & Compressed Size \\ \midrule
\multicolumn{1}{c|}{\multirow{4}{*}{MNIST}}    & \cellcolor{Gray}\textbf{Benign}   & \cellcolor{Gray}\textbf{1.741}         & \cellcolor{Gray}\textbf{1.887}          & \cellcolor{Gray}\textbf{988.89 B}      & \cellcolor{Gray}\textbf{431.40 B}        \\
\multicolumn{1}{c|}{}                          & FGSM~\citeyearpar{goodfellow2014explaining}     & 2.488         & 2.601          & 1690.36 B     & 503.54 B        \\
\multicolumn{1}{c|}{}                          & DeepFool~\citeyearpar{moosavi2015deepfool} & 4.844         & 5.088          & 1654.99 B     & 510.41 B        \\
\multicolumn{1}{c|}{}                          & CW ($L_2$)~\citeyearpar{carlini2016towards}      & 4.094         & 4.301          & 1159.01 B     & 437.27 B        \\ \midrule
\multicolumn{1}{c|}{\multirow{4}{*}{CIFAR-10}} & \cellcolor{Gray}\textbf{Benign}   & \cellcolor{Gray}\textbf{9.595}         & \cellcolor{Gray}\textbf{7.104}       & \cellcolor{Gray}\textbf{1845.98 B}     & \cellcolor{Gray}\textbf{741.36 B}        \\
\multicolumn{1}{c|}{}                          & FGSM~\citeyearpar{goodfellow2014explaining}     & 9.937         & 7.710          & 2717.01 B     & 872.40 B        \\
\multicolumn{1}{c|}{}                          & DeepFool~\citeyearpar{moosavi2015deepfool} & 9.675         & 7.147          & 1880.41 B     & 743.02 B        \\
\multicolumn{1}{c|}{}                          & CW ($L_2$)~\citeyearpar{carlini2016towards}      & 9.621         & 7.113          & 1850.54 B     & 741.56 B        \\ \bottomrule
\end{tabular}
\end{small}
\end{table}


\begin{table}
\centering
\caption{Comparison of complexity for benign and adversarial examples on IMDB and Reuters2.}
\label{tab:entropy_text}
\begin{small}
\begin{tabular}{@{}cccccc@{}}
\toprule
Dataset                                        & Examples & Mean Bits & H (BW) & H (bW) & Compressed Size \\ \midrule
\multicolumn{1}{c|}{\multirow{4}{*}{IMDB}}     & \cellcolor{Gray}\textbf{Benign}   & \cellcolor{Gray}\textbf{4.556}     & \cellcolor{Gray}\textbf{0.569}            & \cellcolor{Gray}\textbf{0.99775}         & \cellcolor{Gray}\textbf{2.235 B}         \\
\multicolumn{1}{c|}{}                          & FGSM~\citeyearpar{goodfellow2014explaining}     & 4.671     & 0.584            & 0.99926         & 3.027 B         \\
\multicolumn{1}{c|}{}                          & FGVM~\citeyearpar{miyato2015distributional}     & 4.701     & 0.588            & 0.99944         & 3.481 B         \\
\multicolumn{1}{c|}{}                          & DeepFool~\citeyearpar{moosavi2015deepfool} & 4.632     & 0.580            & 0.99953         & 3.156 B         \\ \midrule
\multicolumn{1}{c|}{\multirow{4}{*}{Reuters2}} & \cellcolor{Gray}\textbf{Benign}   & \cellcolor{Gray}\textbf{4.946}     &\cellcolor{Gray}\textbf{0.618}            & \cellcolor{Gray}\textbf{0.99457}         & \cellcolor{Gray}\textbf{1.934 B}         \\
\multicolumn{1}{c|}{}                          & FGSM~\citeyearpar{goodfellow2014explaining}     & 5.032     & 0.629            & 0.99712         & 3.181 B         \\
\multicolumn{1}{c|}{}                          & FGVM~\citeyearpar{miyato2015distributional}     & 5.035     & 0.629            & 0.99754         & 3.237 B         \\
\multicolumn{1}{c|}{}                          & DeepFool~\citeyearpar{moosavi2015deepfool} & 5.202     & 0.650            & 0.99545         & 3.301 B         \\ \bottomrule
\end{tabular}
\end{small}
\vspace{-2mm}
\end{table}

\subsection{Input Complexity Estimation}

\label{sec:input_complexity}
\begin{wrapfigure}{R}{0.3\textwidth}
\vspace{-6mm}
  \begin{center}
    \includegraphics[width=0.3\textwidth]{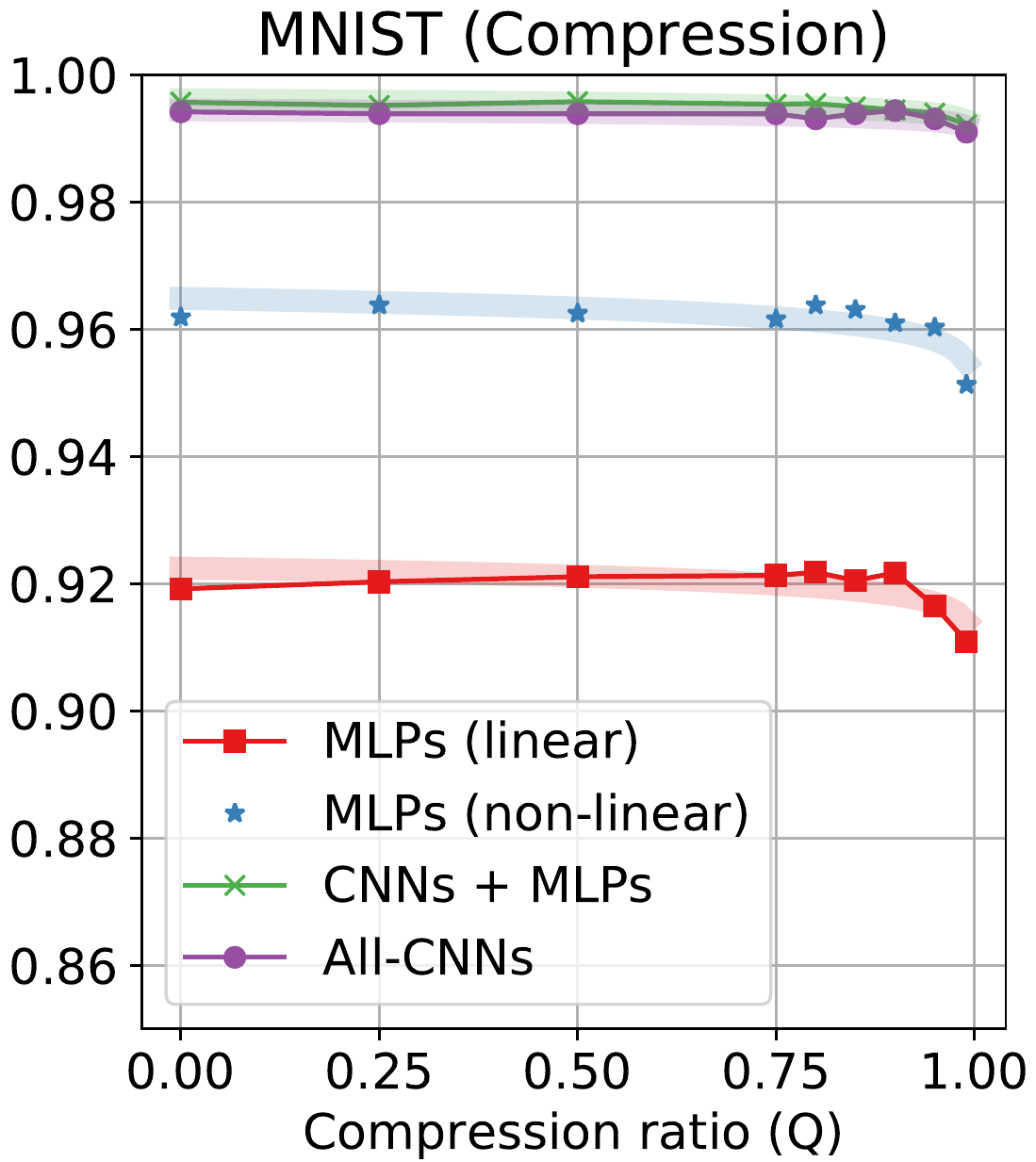}
  \end{center}
  \caption{Classification accuracy as a function of the JPEG compression quality $q$. The shadow curve represents the properly scaled version of the theoretical curve in~\cite{acmmm2018}.}
  \vspace{-12mm}
  \label{fig:lossy_compression}
\end{wrapfigure}
We now investigate if there are possible ways to exploit the higher complexity of adversarial examples to possibly detect adversarial attacks. That is to say, we need a machine-learning independent measure of entropy to evaluate how much benign and adversarial examples differ. For images, we utilized Maximum Likelihood (MLE), Minimax (JVHW)~\citep{JiaoVW14a} and compression estimators for three kind of adversarial examples (FGSM~\citep{goodfellow2014explaining}, DeepFool~\citep{moosavi2015deepfool}, CW ($L_2$)~\citep{carlini2016towards}) on both MNIST and CIFAR-10 dataset. These are four metrics for entropy measurement, all of which indicate higher unpredictability with the value increasing. For compression estimation, prior work~\citep{acmmm2018} has found that an optimal quantification ratio exists for DNNs and appropriate perceptual compression is not harmful. Therefore, we consider such information as redundancy and set the quality scale to 20 following their strategy. We also reproduce the experiments in our settings and obtain the same results shown in Figure~\ref{fig:lossy_compression}.

As shown in Table~\ref{tab:entropy_img}, the benign images have smallest complexity in all of the four metrics, which suggests less entropy (lower complexity) and therefore higher predictability. Similarly, we also design four metrics for text entropy estimation including mean bits per character, byte-wise entropy (BW), bit-wise entropy (bW) and compression size. More specifically, BW and bW are calculated based on the histogram of the bytes or the bits per word. It is worthwhile to note that all of the metrics are measured on adversarial-benign altered word pairs, because adversarial algorithms only modify specific words of the texts. In our evaluations, FGSM, FGVM~\citep{miyato2015distributional} and DeepFool attacks are implemented. From Table~\ref{tab:entropy_text}, we can draw the conclusion that adversarial texts introduce more redundant bits with regards to the target function which results in higher complexity and therefore higher entropy. A reduction of adversarial attacks via entropy measurement is therefore potentially possible for both data types.
\label{sec:output_complexity}
\begin{figure}[htbp!]
\centering 
\begin{subfigure}[b]{.29\textwidth}
    \centering
    \includegraphics[width=\textwidth]{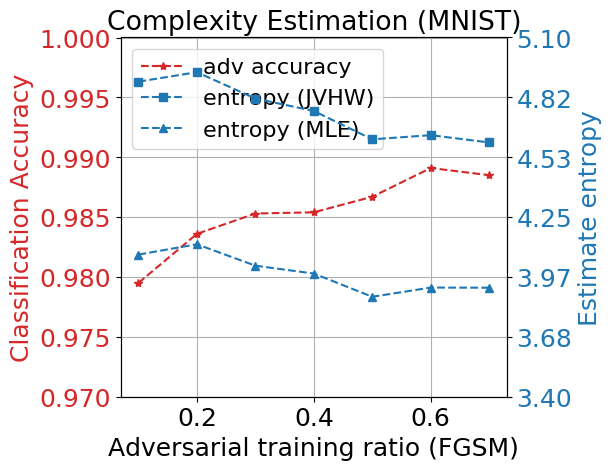}
\end{subfigure}
\begin{subfigure}[b]{.29\textwidth}
    \centering
    \includegraphics[width=\textwidth]{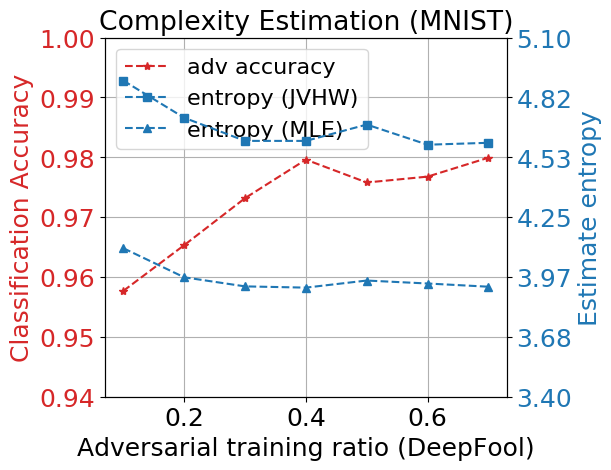}
\end{subfigure}
\begin{subfigure}[b]{.29\textwidth}
    \centering
    \includegraphics[width=\textwidth]{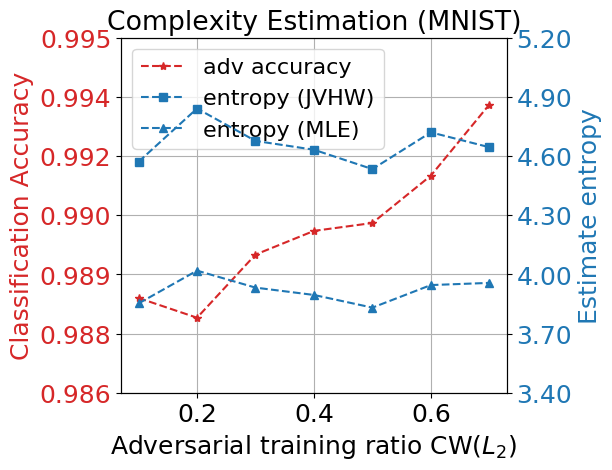}
\end{subfigure}

\begin{subfigure}[b]{.29\textwidth}
    \centering
    \includegraphics[width=\textwidth]{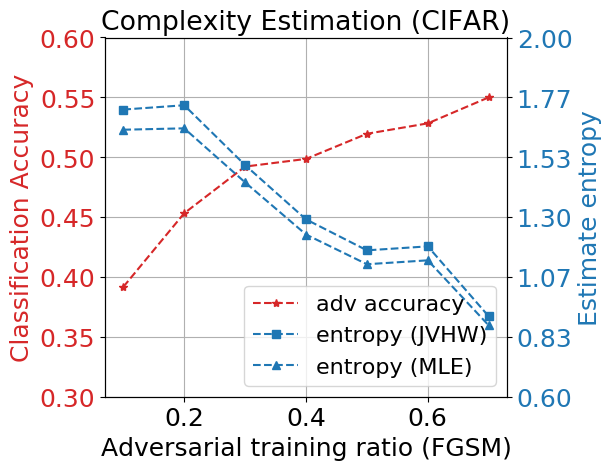}
    \caption{FGSM}
\end{subfigure}
\begin{subfigure}[b]{.29\textwidth}
    \centering
    \includegraphics[width=\textwidth]{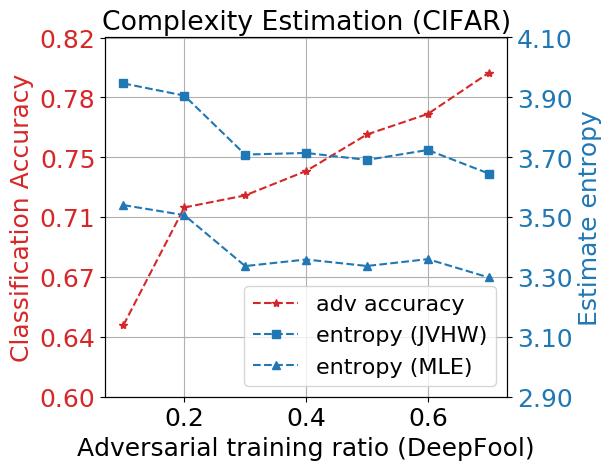}
    \caption{DeepFool}
\end{subfigure}
\begin{subfigure}[b]{.29\textwidth}
    \centering
    \includegraphics[width=\textwidth]{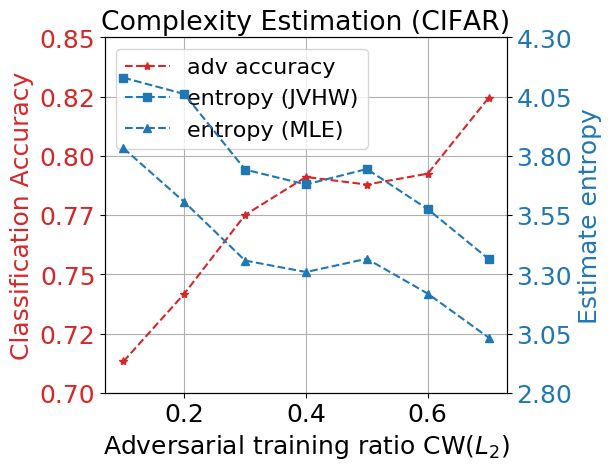}
    \caption{CW($L_2$)}
\end{subfigure}

\caption{Estimated entropy of the output layer under All-CNNs architecture with different robustness. Three adversarial attacks (FGSM, DeepFool, CW) are tested on MNIST and CIFAR classification dataset. Note that only the adversarial examples that successfully fool the networks are calculated.}
\label{fig:output_complexity}
\end{figure}

\subsection{Output Complexity Estimation}
Inspired by Theorem~\ref{thm:necessary}, we investigate the relation between feature redundancy and the robustness of an ML model. We expect that more robust models would employ more succinct features for prediction. To validate this, we design models with different levels of robustness and 
measure the entropy of extracted features (i.e., the input to the final activation layer for decisions). In the experiments, we choose All-CNNs network for it has no fully-connected layer and the last convolutional layer is directly followed by the global average pooling and softmax activation layer, which is convenient for the estimation of entropy. In other words, we can estimate the entropy of the feature maps extracted by the last convolutional layer using perceptual compression and MLE/JVHW estimators. Specifically, we train different models on benign examples and control the ratios of adversarial examples in adversarial re-training period to obtain models with different robustness. In general, a larger ratio of adversarial examples in the training set will lead to a more robust model. The robustness is measured by the test accuracy on adversarial examples. Then we obtain the feature maps on adversarial examples generated by these models and compress them to $q = 20$, following~\cite{acmmm2018}. Finally, we measure the compressed entropy of using MLE and JVHW estimator like Section~\ref{sec:input_complexity}. As illustrated in Figure~\ref{fig:output_complexity}, the estimated entropy (blue dots) decreases as the classification accuracy (red dots) increases for all the three adversarial attacks (FGSM, DeepFool, CW) and the two datasets (MNIST, CIFAR), which means that the redundancy of last-layer feature maps is lower when the models become more robust. Surprisingly, adding adversarial examples into the training set serves as an implicit regularizer for feature redundancy.

\section{Conclusion and Recommendations}
\label{sec:conclusion}
\gerald{reformulated the first two sentences based on ruoxis input}
Our theoretical and empirical results presented in this paper consistently show that adversarial examples are enabled by irrelevant input that the networks are not trained to suppress. In fact, a single bit of redundancy can be exploited to cause the ML models to make arbitrary mistakes. Moreover, redundancy exploited against one model can also affect the decision of another model trained on the same data as that other model learned to only cope with the same amount of redundancy (transferability-based attack). Unfortunately, unlike the academic example in Section~\ref{sec:boolmodel}, we almost never know how many variables we actually need. For image classification, for example, the current assumption is that each pixel serves as input and it is well known that this is feeding the network redundant information e.g., nobody would assume that the upper-most left-most pixel contributes to an object recognition result when the object is usually centered in the image. 

Nevertheless, the highest priority actionable recommendation has to be to reduce redundancies. Before deep learning, manually-crafted features reduced redundancies assumed by humans before the data entered the ML system. This practice has been abandoned with the introduction of deep learning, explaining the temporal correlation with the discovery of adversarial examples. Short of going back to manual feature extraction, automatic techniques can be used to reduce redundancy. Obviously, adaptive techniques, like auto encoders, will be susceptible to their own adversarial attacks. However, consistent with our experiments in Section~\ref{sec:input_complexity}, and dependent on the input domain, we recommend to use lossy compression.   Similar results using quantization have been reported for MP3 and audio compression~\citep{acmmm2018} as well as molecular dynamics~\citep{shuai2018}. In general, we recommend a training procedure where input data is increasingly quantized while training accuracy is measured. The point where the highest quantization is achieved at limited loss in accuracy, is the point where most of the noise and least of the content is lost. This should be the point with least redundancies and therefore the operation point least susceptible to adversarial attacks. In terms of detecting adversarial examples, we showed in Section~\ref{sec:experiments} that estimating the complexity of the input using surrogate methods, such as different compression techniques, can serve as a prefilter to detect adversarial attacks. We will dedicate future work to this topic. Ultimately, however, the only way to practically guarantee adversarial attacks cannot happen is to present every possible input to the machine learner and train to 100\% accuracy, which contradicts the idea of generalization in ML itself. There is no free lunch.

\subsubsection*{Acknowledgments}
This work was partially performed under the auspices of the U.S. Department of Energy by Lawrence Livermore National Laboratory under Contract DE-AC52-07NA27344. It was partially supported by a Lawrence Livermore Laboratory Directed Research \& Development grant (18-ERD-021, ``Explainable AI"). IM release number LLNL-CONF-752187. This work was also supported in part by the Republic of Singapore's National Research Foundation through a grant to the Berkeley Education Alliance for Research in Singapore (BEARS) for the Singapore-Berkeley Building Efficiency and Sustainability in the Tropics (SinBerBEST) Program. BEARS has been established by the University of California, Berkeley as a center for intellectual excellence in research and education in Singapore. Furthermore, this work has also supported by the National Science Foundation under award numbers CNS-1238959, CNS-1238962, CNS-1239054, CNS-1239166, CNS 1514509, and CI-P 1629990. Any findings and conclusions are those of the authors, and do not necessarily reflect the views of the funders. We want to cordially thank Alfredo Metere, Jerome Feldman, Kannan Ramchandran, Bhiksha Raj, and Nathan Mundhenk for their insightful advise.


\bibliography{ref}

\begin{thebibliography}{46}
\providecommand{\natexlab}[1]{#1}
\providecommand{\url}[1]{\texttt{#1}}
\expandafter\ifx\csname urlstyle\endcsname\relax
  \providecommand{\doi}[1]{doi: #1}\else
  \providecommand{\doi}{doi: \begingroup \urlstyle{rm}\Url}\fi

\bibitem[Athalye et~al.(2018)Athalye, Carlini, and
  Wagner]{athalye2018obfuscated}
Anish Athalye, Nicholas Carlini, and David Wagner.
\newblock Obfuscated gradients give a false sense of security: Circumventing
  defenses to adversarial examples.
\newblock \emph{arXiv preprint arXiv:1802.00420}, 2018.

\bibitem[Biggio et~al.(2013)Biggio, Corona, Maiorca, Nelson, {\v{S}}rndi{\'c},
  Laskov, Giacinto, and Roli]{biggio2013evasion}
Battista Biggio, Igino Corona, Davide Maiorca, Blaine Nelson, Nedim
  {\v{S}}rndi{\'c}, Pavel Laskov, Giorgio Giacinto, and Fabio Roli.
\newblock Evasion attacks against machine learning at test time.
\newblock In \emph{Joint European Conference on Machine Learning and Knowledge
  Discovery in Databases}, pp.\  387--402. Springer, 2013.

\bibitem[Cao \& Gong(2017)Cao and Gong]{cao2017mitigating}
Xiaoyu Cao and Neil~Zhenqiang Gong.
\newblock Mitigating evasion attacks to deep neural networks via region-based
  classification.
\newblock \emph{arXiv preprint arXiv:1709.05583}, 2017.

\bibitem[Carlini \& Wagner(2017{\natexlab{a}})Carlini and
  Wagner]{carlini2016towards}
Nicholas Carlini and David Wagner.
\newblock Towards evaluating the robustness of neural networks.
\newblock In \emph{IEEE Symposium on Security and Privacy, 2017},
  2017{\natexlab{a}}.

\bibitem[Carlini \& Wagner(2017{\natexlab{b}})Carlini and
  Wagner]{carlini2017adversarial}
Nicholas Carlini and David Wagner.
\newblock Adversarial examples are not easily detected: Bypassing ten detection
  methods.
\newblock In \emph{Proceedings of the 10th ACM Workshop on Artificial
  Intelligence and Security}, pp.\  3--14. ACM, 2017{\natexlab{b}}.

\bibitem[Carlini \& Wagner(2017{\natexlab{c}})Carlini and
  Wagner]{carlini2017towards}
Nicholas Carlini and David Wagner.
\newblock Towards evaluating the robustness of neural networks.
\newblock In \emph{Security and Privacy (SP), 2017 IEEE Symposium on}, pp.\
  39--57. IEEE, 2017{\natexlab{c}}.

\bibitem[Carlini \& Wagner(2018)Carlini and Wagner]{carlini2018audio}
Nicholas Carlini and David Wagner.
\newblock Audio adversarial examples: Targeted attacks on speech-to-text.
\newblock \emph{arXiv preprint arXiv:1801.01944}, 2018.

\bibitem[Chen et~al.(2018)Chen, Wang, Li, Lu, Luo, Xue, and Wang]{Chen2018}
Yiping Chen, Jingkang Wang, Jonathan Li, Cewu Lu, Zhipeng Luo, Han Xue, and
  Cheng Wang.
\newblock Lidar-video driving dataset: Learning driving policies effectively.
\newblock In \emph{The IEEE Conference on Computer Vision and Pattern
  Recognition (CVPR)}, June 2018.

\bibitem[Ciregan et~al.(2012)Ciregan, Meier, and Schmidhuber]{ciregan2012multi}
Dan Ciregan, Ueli Meier, and J{\"u}rgen Schmidhuber.
\newblock Multi-column deep neural networks for image classification.
\newblock In \emph{Computer vision and pattern recognition (CVPR), 2012 IEEE
  conference on}, pp.\  3642--3649. IEEE, 2012.

\bibitem[Ebrahimi et~al.(2017)Ebrahimi, Rao, Lowd, and
  Dou]{ebrahimi2017hotflip}
Javid Ebrahimi, Anyi Rao, Daniel Lowd, and Dejing Dou.
\newblock Hotflip: White-box adversarial examples for nlp.
\newblock \emph{arXiv preprint arXiv:1712.06751}, 2017.

\bibitem[Evtimov et~al.(2017)Evtimov, Eykholt, Fernandes, Kohno, Li, Prakash,
  Rahmati, and Song]{evtimov2017robust}
Ivan Evtimov, Kevin Eykholt, Earlence Fernandes, Tadayoshi Kohno, Bo~Li, Atul
  Prakash, Amir Rahmati, and Dawn Song.
\newblock Robust physical-world attacks on machine learning models.
\newblock \emph{arXiv preprint arXiv:1707.08945}, 2017.

\bibitem[Fawzi \& Frossard(2015)Fawzi and Frossard]{fawzi2015manitest}
Alhussein Fawzi and Pascal Frossard.
\newblock Manitest: Are classifiers really invariant?
\newblock \emph{arXiv preprint arXiv:1507.06535}, 2015.

\bibitem[Feinman et~al.(2017)Feinman, Curtin, Shintre, and
  Gardner]{feinman2017detecting}
Reuben Feinman, Ryan~R Curtin, Saurabh Shintre, and Andrew~B Gardner.
\newblock Detecting adversarial samples from artifacts.
\newblock \emph{arXiv preprint arXiv:1703.00410}, 2017.

\bibitem[Feynman et~al.(2000)Feynman, Hey, and Allen]{feynman88}
Richard~Phillips Feynman, Anthony~J Hey, and Robin~W Allen.
\newblock \emph{Feynman lectures on computation}.
\newblock Perseus Books, 2000.

\bibitem[Friedland \& Krell(2017)Friedland and Krell]{friedlandkrell2017}
Gerald Friedland and Mario Krell.
\newblock A capacity scaling law for artificial neural networks.
\newblock \emph{arXiv preprint arXiv:1708.06019}, 2017.

\bibitem[Friedland et~al.(2018{\natexlab{a}})Friedland, Metere, and
  Krell]{friedland2018}
Gerald Friedland, Alfredo Metere, and Mario Krell.
\newblock A practical approach to sizing neural networks.
\newblock \emph{arXiv preprint arXiv:1810.02328}, 2018{\natexlab{a}}.

\bibitem[Friedland et~al.(2018{\natexlab{b}})Friedland, Wang, Jia, and
  Li]{acmmm2018}
Gerald Friedland, Jingkang Wang, Ruoxi Jia, and Bo~Li.
\newblock The helmholtz method: Using perceptual compression to reduce machine
  learning complexity.
\newblock \emph{arXiv preprint arXiv:1807.10569}, 2018{\natexlab{b}}.

\bibitem[Gilmer et~al.(2018)Gilmer, Metz, Faghri, Schoenholz, Raghu,
  Wattenberg, and Goodfellow]{gilmer2018adversarial}
Justin Gilmer, Luke Metz, Fartash Faghri, Samuel~S Schoenholz, Maithra Raghu,
  Martin Wattenberg, and Ian Goodfellow.
\newblock Adversarial spheres.
\newblock \emph{arXiv preprint arXiv:1801.02774}, 2018.

\bibitem[Goodfellow et~al.(2015)Goodfellow, Shlens, and
  Szegedy]{goodfellow2014explaining}
Ian~J Goodfellow, Jonathon Shlens, and Christian Szegedy.
\newblock Explaining and harnessing adversarial examples.
\newblock In \emph{International Conference on Learning Representations}, 2015.

\bibitem[He et~al.(2018)He, Li, and Song]{he2018decision}
Warren He, Bo~Li, and Dawn Song.
\newblock Decision boundary analysis of adversarial examples.
\newblock In \emph{ICLR-International Conference on Learning Representations},
  2018.

\bibitem[Hinton et~al.(2012)Hinton, Deng, Yu, Dahl, Mohamed, Jaitly, Senior,
  Vanhoucke, Nguyen, Sainath, et~al.]{hinton2012deep}
Geoffrey Hinton, Li~Deng, Dong Yu, George~E Dahl, Abdel-rahman Mohamed, Navdeep
  Jaitly, Andrew Senior, Vincent Vanhoucke, Patrick Nguyen, Tara~N Sainath,
  et~al.
\newblock Deep neural networks for acoustic modeling in speech recognition: The
  shared views of four research groups.
\newblock \emph{IEEE Signal Processing Magazine}, 29\penalty0 (6):\penalty0
  82--97, 2012.

\bibitem[Huang et~al.(2017)Huang, Papernot, Goodfellow, Duan, and
  Abbeel]{huang2017adversarial}
Sandy Huang, Nicolas Papernot, Ian Goodfellow, Yan Duan, and Pieter Abbeel.
\newblock Adversarial attacks on neural network policies.
\newblock \emph{arXiv preprint arXiv:1702.02284}, 2017.

\bibitem[Jiao et~al.(2014)Jiao, Venkat, and Weissman]{JiaoVW14a}
Jiantao Jiao, Kartik Venkat, and Tsachy Weissman.
\newblock Order-optimal estimation of functionals of discrete distributions.
\newblock \emph{CoRR}, abs/1406.6956, 2014.
\newblock URL \url{http://arxiv.org/abs/1406.6956}.

\bibitem[Kanbak(2017)]{kanbak2017measuring}
Can Kanbak.
\newblock Measuring robustness of classifiers to geometric transformations.
\newblock Technical report, 2017.

\bibitem[Kos \& Song(2017)Kos and Song]{kos2017delving}
Jernej Kos and Dawn Song.
\newblock Delving into adversarial attacks on deep policies.
\newblock \emph{arXiv preprint arXiv:1705.06452}, 2017.

\bibitem[Kurakin et~al.(2016)Kurakin, Goodfellow, and
  Bengio]{kurakin2016adversarial}
Alexey Kurakin, Ian Goodfellow, and Samy Bengio.
\newblock Adversarial examples in the physical world.
\newblock \emph{arXiv preprint arXiv:1607.02533}, 2016.

\bibitem[Li \& Li(2017)Li and Li]{li2017adversarial}
Xin Li and Fuxin Li.
\newblock Adversarial examples detection in deep networks with convolutional
  filter statistics.
\newblock In \emph{ICCV}, pp.\  5775--5783, 2017.

\bibitem[Liu(2018)]{shuai2018}
Shuai Liu.
\newblock \emph{Compression and Analysis of Molecular Dynamics Data}.
\newblock University of California, Berkeley, 2018.

\bibitem[Liu et~al.(2017)Liu, Chen, Liu, and Song]{liu2016delving}
Yanpei Liu, Xinyun Chen, Chang Liu, and Dawn Song.
\newblock Delving into transferable adversarial examples and black-box attacks.
\newblock In \emph{ICLR}, 2017.

\bibitem[Liu et~al.(2018)Liu, Liu, Liu, Wang, and Wen]{liu2018feature}
Zihao Liu, Qi~Liu, Tao Liu, Yanzhi Wang, and Wujie Wen.
\newblock Feature distillation: Dnn-oriented jpeg compression against
  adversarial examples.
\newblock \emph{arXiv preprint arXiv:1803.05787}, 2018.

\bibitem[Ma et~al.(2018)Ma, Li, Wang, Erfani, Wijewickrema, Houle, Schoenebeck,
  Song, and Bailey]{ma2018characterizing}
Xingjun Ma, Bo~Li, Yisen Wang, Sarah~M Erfani, Sudanthi Wijewickrema, Michael~E
  Houle, Grant Schoenebeck, Dawn Song, and James Bailey.
\newblock Characterizing adversarial subspaces using local intrinsic
  dimensionality.
\newblock \emph{arXiv preprint arXiv:1801.02613}, 2018.

\bibitem[MacKay(2003)]{mackay2002}
David~JC MacKay.
\newblock \emph{Information theory, inference and learning algorithms}.
\newblock Cambridge university press, 2003.

\bibitem[Miyato et~al.(2015)Miyato, Maeda, Koyama, Nakae, and
  Ishii]{miyato2015distributional}
Takeru Miyato, Shin-ichi Maeda, Masanori Koyama, Ken Nakae, and Shin Ishii.
\newblock Distributional smoothing with virtual adversarial training.
\newblock \emph{stat}, 1050:\penalty0 25, 2015.

\bibitem[Moosavi-Dezfooli et~al.(2015)Moosavi-Dezfooli, Fawzi, and
  Frossard]{moosavi2015deepfool}
Seyed-Mohsen Moosavi-Dezfooli, Alhussein Fawzi, and Pascal Frossard.
\newblock Deepfool: a simple and accurate method to fool deep neural networks.
\newblock \emph{arXiv preprint arXiv:1511.04599}, 2015.

\bibitem[Moosavi-Dezfooli et~al.(2016)Moosavi-Dezfooli, Fawzi, Fawzi, and
  Frossard]{moosavi2016universal}
Seyed-Mohsen Moosavi-Dezfooli, Alhussein Fawzi, Omar Fawzi, and Pascal
  Frossard.
\newblock Universal adversarial perturbations.
\newblock \emph{arXiv preprint arXiv:1610.08401}, 2016.

\bibitem[Mopuri et~al.(2017)Mopuri, Garg, and Babu]{mopuri2017fast}
Konda~Reddy Mopuri, Utsav Garg, and R~Venkatesh Babu.
\newblock Fast feature fool: A data independent approach to universal
  adversarial perturbations.
\newblock \emph{arXiv preprint arXiv:1707.05572}, 2017.

\bibitem[Papernot et~al.(2016{\natexlab{a}})Papernot, McDaniel, and
  Goodfellow]{papernot2016transferability}
Nicolas Papernot, Patrick McDaniel, and Ian Goodfellow.
\newblock Transferability in machine learning: from phenomena to black-box
  attacks using adversarial samples.
\newblock \emph{arXiv preprint arXiv:1605.07277}, 2016{\natexlab{a}}.

\bibitem[Papernot et~al.(2016{\natexlab{b}})Papernot, McDaniel, Jha,
  Fredrikson, Celik, and Swami]{papernot2016limitations}
Nicolas Papernot, Patrick McDaniel, Somesh Jha, Matt Fredrikson, Z~Berkay
  Celik, and Ananthram Swami.
\newblock The limitations of deep learning in adversarial settings.
\newblock In \emph{2016 IEEE European Symposium on Security and Privacy
  (EuroS\&P)}, pp.\  372--387. IEEE, 2016{\natexlab{b}}.

\bibitem[Phillips(1990)]{phillips1990neural}
Dennis~P Phillips.
\newblock Neural representation of sound amplitude in the auditory cortex:
  effects of noise masking.
\newblock \emph{Behavioural brain research}, 37\penalty0 (3):\penalty0
  197--214, 1990.

\bibitem[Ponomarenko et~al.(2007)Ponomarenko, Silvestri, Egiazarian, Carli,
  Astola, and Lukin]{ponomarenko2007between}
Nikolay Ponomarenko, Flavia Silvestri, Karen Egiazarian, Marco Carli, Jaakko
  Astola, and Vladimir Lukin.
\newblock On between-coefficient contrast masking of dct basis functions.
\newblock In \emph{Proceedings of the third international workshop on video
  processing and quality metrics}, volume~4, 2007.

\bibitem[Rojas(2013)]{Rojas1992}
Ra{\'u}l Rojas.
\newblock \emph{Neural networks: a systematic introduction}.
\newblock Springer Science \& Business Media, 2013.

\bibitem[Sallab et~al.(2017)Sallab, Abdou, Perot, and Yogamani]{sallab2017deep}
Ahmad~EL Sallab, Mohammed Abdou, Etienne Perot, and Senthil Yogamani.
\newblock Deep reinforcement learning framework for autonomous driving.
\newblock \emph{Electronic Imaging}, 2017\penalty0 (19):\penalty0 70--76, 2017.

\bibitem[Szegedy et~al.(2014)Szegedy, Zaremba, Sutskever, Bruna, Erhan,
  Goodfellow, and Fergus]{szegedy2014intriguing}
Christian Szegedy, Wojciech Zaremba, Ilya Sutskever, Joan Bruna, Dumitru Erhan,
  Ian Goodfellow, and Rob Fergus.
\newblock Intriguing properties of neural networks.
\newblock In \emph{International Conference on Learning Representations}, 2014.

\bibitem[Tram{\`e}r et~al.(2017)Tram{\`e}r, Papernot, Goodfellow, Boneh, and
  McDaniel]{tramer2017space}
Florian Tram{\`e}r, Nicolas Papernot, Ian Goodfellow, Dan Boneh, and Patrick
  McDaniel.
\newblock The space of transferable adversarial examples.
\newblock \emph{arXiv preprint arXiv:1704.03453}, 2017.

\bibitem[Wang et~al.(2016)Wang, Gao, and Qi]{wang2016theoretical}
Beilun Wang, Ji~Gao, and Yanjun Qi.
\newblock A theoretical framework for robustness of (deep) classifiers against
  adversarial examples.
\newblock \emph{arXiv preprint arXiv:1612.00334}, 2016.

\bibitem[Warm et~al.(1997)Warm, Wittmer, and Noll]{warm1997apparatus}
Berndt Warm, Detlev Wittmer, and Matthias Noll.
\newblock Apparatus for defending against an attacking missile, February~4
  1997.
\newblock US Patent 5,600,434.

\end{thebibliography}
\bibliographystyle{iclr2019_conference}

\newpage
\appendix
\section{Proof of Theorem~\ref{thm:necessary}}

\begin{proof}
Let $\mathcal{X}$ be the set of admissible data points and $\mathcal{X}'$ denote the set of adversarial examples,We prove this theorem by constructing a sufficient statistic $T'(X)$ that has lower entropy than $T(X)$. 
Consider
\begin{align}
\label{eqn:feature}
    T'(x') = \left\{\begin{array}{l}
         T(x') \text{  if $x'\in \mathcal{X}\setminus \mathcal{X}'$}  \\
         T(x)  \text{  if $x'\in \mathcal{X}'$ }
    \end{array}\right .
\end{align}

where $x$ is an arbitrary benign example in the data space. Then, for all $x'\in \mathcal{X}'$, $g(T(x))\neq g(T(x'))$. It follows that $T(x)\neq T(x'), \forall x'\in \mathcal{X}'$. On the other hand, $T(x)=T'(x)$ by construction. 


Let the probability density of $T(X)$ be denoted by $p(t)$, where $t\in T(\mathcal{X})$, and the probability density of $T'(X)$ be denoted by $q(t)$ where $t\in T(\mathcal{X}\setminus \mathcal{X}')$. Then, $q(t) = p(t) + w(t)$ for $t\in T(\mathcal{X}\setminus \mathcal{X}')$, where $w(t)$ corresponds to the part of benign example probability that is formed by enforcing an originally adversarial example' feature to be equal to the feature of an arbitrary benign example according to (\ref{eqn:feature}). Furthermore, $\sum_{t\in T(\mathcal{X}\setminus \mathcal{X}')} w(t) = \sum_{t \in T(\mathcal{X}')} p(t)$. We now compare the entropy of $T(X)$ and $T'(X)$:
\begin{align}
    &H(T(X)) - H(T'(X)) \\
    &= -\sum_{t\in T(\mathcal{X}')} p(t)\log p(t) - \sum_{t\in T(\mathcal{X}\setminus \mathcal{X}')} p(t)\log p(t) + \sum_{t\in T(\mathcal{X}\setminus \mathcal{X}')} (p(t)+w(t))\log (p(t)+w(t))\\
    &=  \underbrace{-\sum_{t\in T(\mathcal{X}')} p(t)\log p(t)}_{U_1} + \underbrace{\sum_{t\in T(\mathcal{X}\setminus \mathcal{X}')} (p(t)+w(t))\log (p(t)+w(t)) - p(t)\log p(t)}_{U_2}
\end{align}

It is evident that $U_1\geq 0$. Note that for any $p(t)$, there always exists a configuration of $w(t)$ such that $U_2\geq 0$. For instance, let $t^* = arg\max_{t\in T(\mathcal{X}\setminus \mathcal{X}')} p(t)$. Then, we can let $w(t^*) = \sum_{t \in T(\mathcal{X}')} p(t)$ and $w(t)=0$ for $t\neq t^*$. With this configuration of $w(t)$, 
\begin{align}
    U_2 = (p(t^*)+w(t^*))\log ((p(t^*)+w(t^*)) - p(t^*)\log p(t^*)
\end{align}
Due to the fact that $x\log x$ is a monotonically increasing function, $U_2\geq 0$. 

To sum up, both $U_1$ and $U_2$ are non-negative; as a result,
\begin{align}
    H(T(X))> H(T'(X))
\end{align}
Thus, we constructed a sufficient statistic $T'(\cdot)$ that achieves lower entropy than $T(\cdot)$, which, in turn, indicates that $T(X)$ is not a minimal sufficient statistic.
\end{proof}

\section{Supplementary Experimental Results}



\begin{figure}[H]
  \centering
  \begin{subfigure}[b]{0.4\textwidth}
    \includegraphics[width=\textwidth]{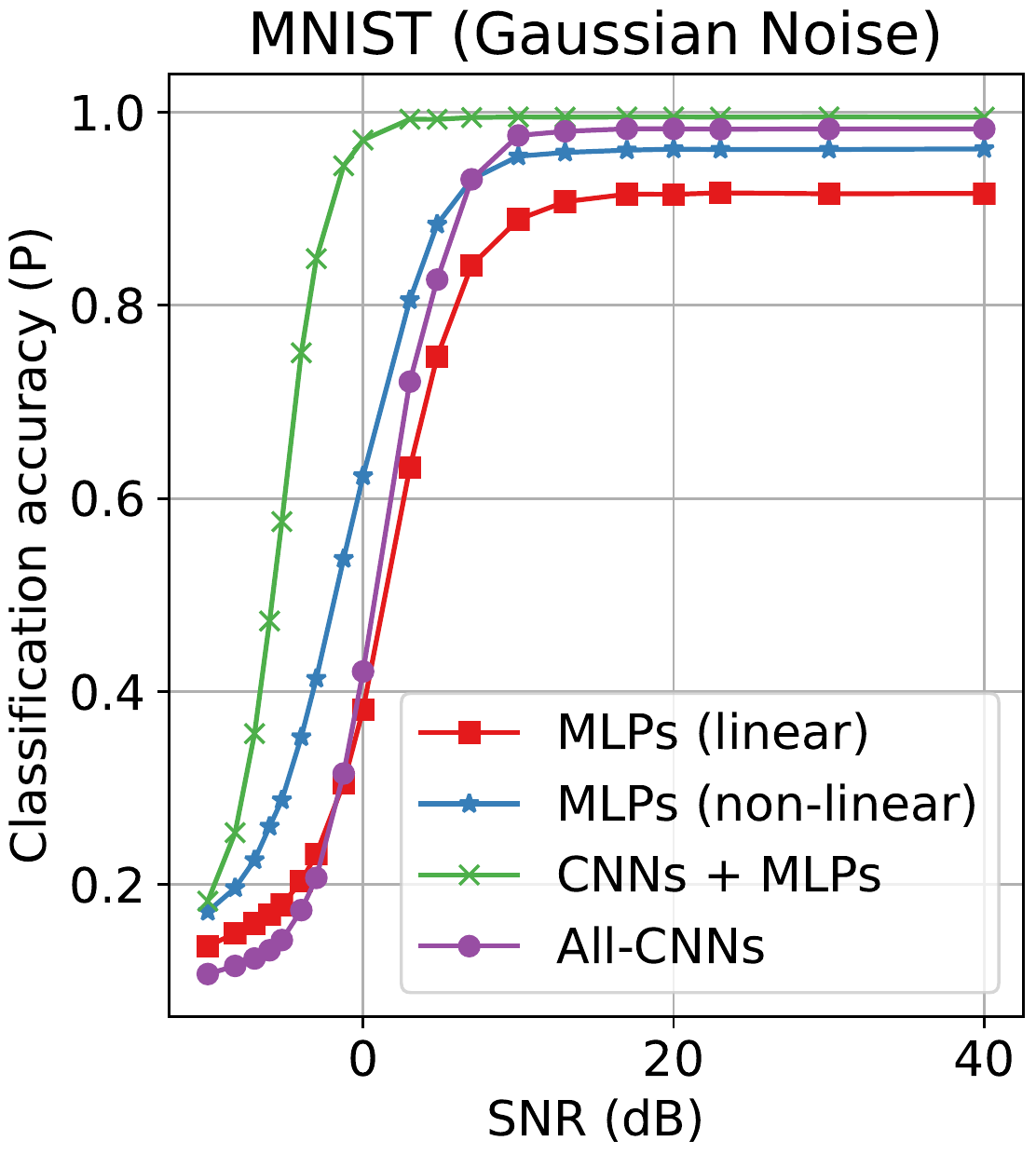}
    \caption{}
    \label{subfig:snr_mnist}
  \end{subfigure}
  \begin{subfigure}[b]{0.38\textwidth}
    \includegraphics[width=\textwidth]{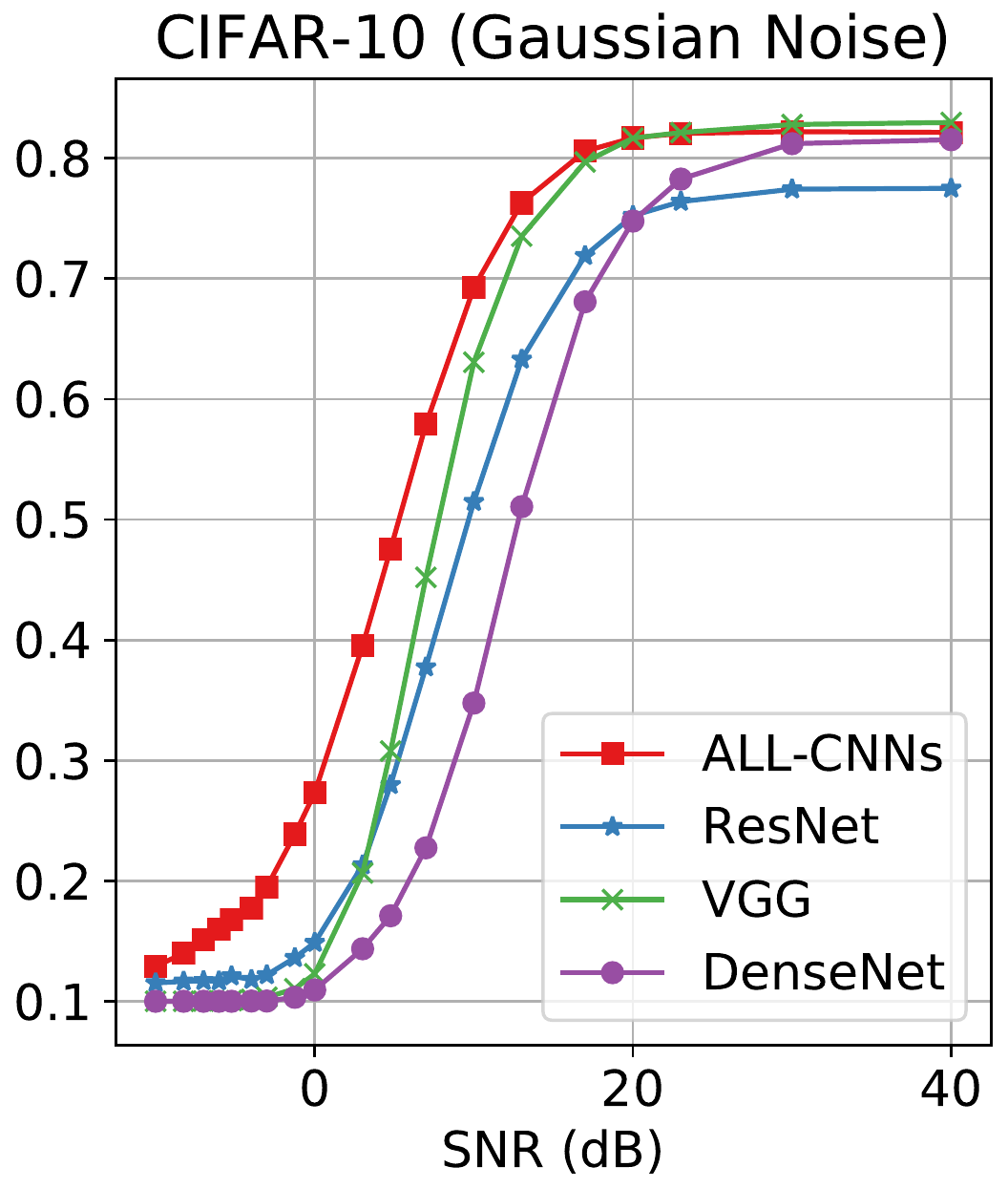}
    \caption{}
    \label{subfig:snr_cifar}
  \end{subfigure}

 \caption{(a, b) Classification accuracy of benign and adversarial examples as a function of signal-to-noise ratio (SNR). (c) Classification accuracy as a function of the JPEG compression quality $q$. The shadow curve represents the properly scaled version of the theoretical curve in~\cite{acmmm2018}.}
\label{fig:snr}
\end{figure}

Apart from the adversarial examples, we also observed the same phenomenon for random noise that redundancy will lead to the failure of DNNs. We tested datasets with different signal-to-noise ratios (SNR), generated by adding Gaussian noise to the real pixels. The SNR is obtained by controlling the variance of the Gaussian distribution. Finally, we derived the testing accuracy on hand-crafted noisy testing data. As shown in Figure~\ref{subfig:snr_mnist} and~\ref{subfig:snr_cifar}, a small amount of random Gaussian noise will add complexity to examples and cause the DNNs to fail. For instance, noisy input sets with one tenth the signal strength of the benign examples result in only $34.3\%$ test accuracy for DenseNet on CIFAR-10. This indeed indicates, and is consistent with related work, that small amounts of noise can practically fool ML models in general. 

\end{document}